\documentclass[conference]{IEEEtran}
\IEEEoverridecommandlockouts
\usepackage{cite}
\usepackage{amsmath,amssymb,amsfonts}
\usepackage{algorithmic}
\usepackage{graphicx}
\usepackage{textcomp}
\usepackage{subcaption}
\usepackage{xcolor}
\def\BibTeX{{\rm B\kern-.05em{\sc i\kern-.025em b}\kern-.08em
    T\kern-.1667em\lower.7ex\hbox{E}\kern-.125emX}}
\begin{document}

\title{\Huge Rehabilitation Exercise Repetition Segmentation and Counting using Skeletal Body Joints\\



}

\author
{
\IEEEauthorblockN{
Ali Abedi\textsuperscript{1},
Paritosh Bisht\textsuperscript{2},
Riddhi Chatterjee\textsuperscript{2},
Rachit Agrawal\textsuperscript{2},
Vyom Sharma\textsuperscript{2},\\
Dinesh Babu Jayagopi\textsuperscript{2},
Shehroz S. Khan\textsuperscript{1,3},
}

\IEEEauthorblockA{
\textsuperscript{1}KITE Research Institute– Toronto Rehabilitation Institute, University Health Network, Canada\\
\textsuperscript{2}Multimodal Perception Lab, International Institute of Information Technology Bangalor, India\\
\textsuperscript{3}Institute of Biomedical Engineering, University of Toronto, Canada\\
    \{ali.abedi, shehroz.khan\}@uhn.ca,
    \{paritosh.bisht, riddhi.chatterjee, rachit.agrawal, vyom.sharma, jdinesh\}@iiitb.ac.in
    }
}

\maketitle

\begin{abstract}
Physical exercise is an essential component of rehabilitation programs that improve quality of life and reduce mortality and re-hospitalization rates. In AI-driven virtual rehabilitation programs, patients complete their exercises independently at home, while AI algorithms analyze the exercise data to provide feedback to patients and report their progress to clinicians. To analyze exercise data, the first step is to segment it into consecutive repetitions. There has been a significant amount of research performed on segmenting and counting the repetitive activities of healthy individuals using raw video data, which raises concerns regarding privacy and is computationally intensive. Previous research on patients' rehabilitation exercise segmentation relied on data collected by multiple wearable sensors, which are difficult to use at home by rehabilitation patients. Compared to healthy individuals, segmenting and counting exercise repetitions in patients is more challenging because of the irregular repetition duration and the variation between repetitions. This paper presents a novel approach for segmenting and counting the repetitions of rehabilitation exercises performed by patients, based on their skeletal body joints. Skeletal body joints can be acquired through depth cameras or computer vision techniques applied to RGB videos of patients. Various sequential neural networks, including many-to-many models (with binary sequence output and density map output) and many-to-one models (with a single output), are designed to analyze the sequences of skeletal body joints and perform repetition segmentation and counting. Extensive experiments on three publicly available rehabilitation exercise datasets, KIMORE, UI-PRMD, and IntelliRehabDS, demonstrate the superiority of the proposed method compared to previous methods. The proposed method enables accurate exercise analysis while preserving privacy, facilitating the effective delivery of virtual rehabilitation programs.
\end{abstract}


\section{Introduction}
Referral of patients to rehabilitation programs following a stroke, cardiac event, or injury is a common practice aimed at improving patients' quality of life and reducing re-hospitalization and death rates \cite{dibben2023exercise}. Central to these programs are regular and repetitive exercises that enable patients to regain mobility and strength \cite{dibben2023exercise}. Recently, Artificial Intelligence (AI)-driven virtual rehabilitation has emerged as a promising approach to delivering rehabilitation programs remotely to patients in their homes \cite{ferreira2023usage}. This approach involves the use of various sensors to capture patients' movements and the use of AI algorithms to analyze patients' movements during exercise \cite{ferreira2023usage,sangani2020real}. The analysis results can be used to provide patients with feedback on the quality or completion of their exercises \cite{sangani2020real,fernandez2018virtualgym}. Additionally, clinicians can also use the analysis results to monitor patients' progress and take appropriate interventions.

In rehabilitation programs, patients are typically prescribed specific exercises with designated numbers of sets and repetitions \cite{liao2020review,dibben2023exercise,capecci2019kimore,vakanski2018data,miron2021intellirehabds}. Evaluating exercise performance relies on objective criteria such as compliance with the prescribed number of sets and repetitions, repeating exercises in a constant manner, proper technique and quality of movements, and correct posture of various body parts \cite{capecci2019kimore,liao2020review}. Therefore, repetition (temporal) segmentation, the process of dividing a continuous sequence of movement data into individual repetitions, is the first step in an AI-driven exercise evaluation pipeline \cite{liao2020review}. Exercise repetition counting can either be derived from the segmentation process or executed as a separate task.

A variety of data modalities were used for repetition segmentation and counting \cite{lin2016movement}, including Inertial Measurement Unit (IMU) sensor data \cite{bevilacqua2019rehabilitation,brennan2020segmentation,lin2018classification,lin2014human,lin2013online,bevilacqua2020combining,soro2019recognition}, video data \cite{hu2022transrac,chunglong,zhang2021repetitive}, and skeletal body joints \cite{hsu2021invariant}. Existing algorithms for segmenting human movement can be divided into unsupervised and supervised algorithms \cite{lin2016movement}. Unsupervised algorithms, which do not require labeled data to develop, include thresholding, template matching, and exemplar-based approaches \cite{sarsfield2019segmentation,lin2016movement}. Supervised algorithms requiring labeled data to be trained on include Support Vector Machines (SVM) \cite{lin2018classification}, Hidden Markov Model (HMM) \cite{lin2013online}, Convolutional Neural Networks (CNNs) \cite{brennan2020segmentation}, and the combination of CNNs and Finite State Machines (FSMs) \cite{bevilacqua2019rehabilitation}. The existing works on repetition counting \cite{zhang2021repetitive,hsu2021invariant,yu2021deep,soro2019recognition} are not capable of segmenting movement data into individual repetitions, that is, they are not capable of determining the start and end timestamps of individual repetitions. Video-based approaches are not privacy-preserving and are computationally prohibitive \cite{hu2022transrac,chunglong,zhang2021repetitive}. The IMU-based approaches require wearing multiple IMU sensors while exercising \cite{bevilacqua2019rehabilitation,brennan2020segmentation,lin2018classification,lin2014human,lin2013online,bevilacqua2020combining,soro2019recognition}, which is challenging in the real world, i.e., in rehabilitation patients' homes. The previous works on healthy individuals \cite{hu2022transrac,chunglong,zhang2021repetitive,hsu2021invariant} cannot be directly used on rehabilitation patients due to the fact that segmenting and counting exercise repetitions is more challenging in patients due to irregularities in the duration and completion of exercises resulting from their respective impairments \cite{capecci2019kimore,liao2020review}.

This paper presents novel methods for rehabilitation exercise repetition segmentation and counting from the skeletal body joints of patients. Various many-to-many and many-to-one deep sequential neural network architectures, including Long Short Term Memory (LSTM) and a combination of LSTM and CNN, are designed to analyze the sequences of skeletal body joints and perform repetition segmentation and counting. Our primary contributions are as follows:
\begin{itemize}
    \item This is the first work on rehabilitation exercise repetition segmentation using skeletal body joints collected by depth cameras or extracted from RGB video using advanced computer-vision techniques.
    \item We developed neural network architectures capable of analyzing sequences of body joints, i.e., multivariate time series.
    \item We conducted extensive experiments on three publicly available rehabilitation exercise datasets and demonstrated the effectiveness of the proposed method compared to the previous methods.
\end{itemize}

As a point of clarification, the purpose of this paper is not rehabilitation exercise recognition/classification nor rehabilitation exercise quality/correctness assessment. Specifically, this paper focuses on the temporal (not spatial) segmentation of rehabilitation exercises into individual repetitions and counting the number of repetitions.


\section{Related Work}
\label{sec:related_work}
This section reviews existing research on repetitive action segmentation and counting, with an emphasis on rehabilitation exercise segmentation and counting. The literature review is organized according to the data modality used for segmentation and counting.

\subsection{Rehabilitation Exercise Segmentation from IMU sensors}
\label{sec:related_work_imu}
Lin et al. \cite{lin2018classification} proposed an approach for segmenting rehabilitation exercises into individual repetitions using the data collected by IMU wearable sensors worn on the hip, knee, and ankle. Segmentation was defined as a binary classification problem in which movement data at consecutive timestamps are classified into segment points or non-segment points. To perform segmentation, after several steps of preprocessing, including filtering, down-sampling, and windowing, the IMU signal is classified into segment or non-segment classes by an SVM. Lin et al. \cite{lin2013online} proposed a two-stage approach in which first, segment point candidates are identified by analyzing the velocity features extracted from the IMU signal. Then, HMMs are used to recognize segment locations from segment point candidates. The method was evaluated on three publicly available IMU datasets and achieved high accuracy. Brennan et al. \cite{bevilacqua2019rehabilitation,brennan2020segmentation} proposed a two-stage approach in which first, a CNN classifies sliding windows of the IMU signals into either "dynamic" or "dormant" classes. The classification results are then streamed into an FSM that keeps track of the classes of consecutive windows and outputs the starting and ending points of repetitions. They achieved high accuracy in shoulder and knee exercises. Bevilacqua et al. \cite{bevilacqua2020combining} proposed a joint rehabilitation exercise motion primitive segmentation and classification using a mixture of LSTMs and boosting aggregation. Their method was evaluated on accelerometer and gyroscope data and achieved high exercise primitive classification accuracy.

\subsection{Rehabilitation Exercise Repetition Counting from Skeletal Body Joints}
\label{sec:related_work_joints}
Hsu et al. \cite{hsu2021invariant} proposed a rehabilitation exercise repetition counting using Skeletal body joints data. Initially, the pairwise cosine similarity of the skeleton time series data is calculated. A spectrogram is then constructed based on the pairwise cosine similarity. A repetition count is then calculated by integrating from the spectrogram. The method was evaluated on the UI-PRMD rehabilitation exercise dataset \cite{vakanski2018data} and the MM-Fit fitness exercise dataset \cite{stromback2020mm} and achieved low Mean Absolute Error (MAE).

\subsection{Repetitive Action Segmentation from Videos}
\label{sec:related_work_video}
Hu et al. \cite{hu2022transrac} introduced a large-scale repetitive action counting dataset, named RepCount, containing 1451 videos with about 20000 annotations of the start and end of repetitions. The dataset contains in-the-wild videos of healthy individuals exercising. They have also proposed a deep neural network architecture, named TransRAC \cite{hu2022transrac}, for repetition counting that was trained and evaluated on the RepCount dataset. In TransRAC, with step sizes of 1, 2, and 4, multi-scale video sequences are generated from the input video. After extracting features from the multi-scale video sequences by an encoder neural network, temporal correlations are calculated between the extracted features, and correlation matrices are created. The concatenation of the correlation matrices is input to a transformer to output a density map as a prediction. The ground-truth density maps are generated by approximating a Gaussian distribution between the start and end of each repetition \cite{hu2022transrac}. To overcome a major limitation of TransRAC, its inability to handle long videos, Chung et al. \cite{chunglong} proposed a video transformer equipped with class token distillation and marginally improved the repetition counting results of TransRAC. Zhang et al. \cite{zhang2021repetitive} proposed an approach for repetition counting from videos incorporating the corresponding sound of the video. An S3D-based architecture was used for repetition counting from video, while a ResNet-18-based neural network was used for repetition counting from audio. The results on an audiovisual dataset showed improvements when audio data is added to video data for repetition counting. To learn more about repetitive action segmentation and counting from videos using deep learning algorithms, please refer to \cite{dwibedi2020counting,ferreira2021deep,cheng2023periodic,jacquelin2022periodicity,yu2021deep,zhang2021repetitive}.

The major disadvantage of the IMU-based methods \cite{lin2018classification,lin2013online,brennan2020segmentation,bevilacqua2019rehabilitation,bevilacqua2020combining} is that they require the determination of various parameters, including window sizes and thresholds. Additionally, these methods are capable of analyzing multivariate time series with a small number of variables, corresponding to the number of IMU sensors worn during exercise, but are unable to analyze multivariate time series with high dimensionalities, such as all the joints of the skeletal system. Moreover, it may be infeasible for rehabilitation patients to wear multiple IMU sensors while exercising independently at home. The method proposed by Hsu et al. \cite{hsu2021invariant} was designed for rehabilitation exercise counting and is not capable of segmenting rehabilitation exercises. Exercise temporal segmentation into individual repetitions is required for accurate exercise assessment \cite{liao2020review}. A limitation of video-based methods \cite{hu2022transrac,chunglong,zhang2021repetitive,dwibedi2020counting,ferreira2021deep,cheng2023periodic,jacquelin2022periodicity,yu2021deep} is their complexity and their large number of training parameters, in addition to privacy concerns. There is no previous method for segmenting rehabilitation exercises based on skeletal body joints \cite{liao2020review}. In order to fill the gap in the literature, this paper proposes deep-learning algorithms for segmenting and counting repetitions in rehabilitation exercises performed by patients. The proposed method facilitates exercise assessment and feedback generation for patients in virtual rehabilitation programs without the need for body-worn sensors.

\section{Method}
\label{sec:methodology}
The input to the proposed method is the sequence of skeletal body joints of a patient doing rehabilitation exercises. The output is the segmented sequence into individual exercise repetitions and the total number of repetitions. The sequence of features extracted from the sequence of skeletal body joints is fed to a sequential neural network for exercise repetition segmentation and counting.

The proposed method can work with either a depth camera capable of capturing skeletal body joints or a regular RGB camera. In the latter case, an additional module is required in order to extract skeletal body joints from RGB video frames. Several libraries are available for extracting body joints from videos, including MediaPipe \cite{lugaresi2019mediapipe} and OpenPose \cite{pavllo20193d}.

The sequential neural networks in the proposed method can be provided with three types of data, raw body joints, exercise-specific features extracted from the body joints, and their concatenation. Inspired by Guo and Khan \cite{guo2021exercise} worked on the KIMORE dataset, exercise-specific features are those calculated based on the angles between the body joints that are moving in specific exercises. For instance, for upper-extremity rehabilitation exercises for stroke patients, the shoulder-wrist angle is an exercise-specific feature \cite{guo2021exercise,capecci2019kimore}. The body joints or features, extracted from each frame of the input data, are provided for each timestamp of a sequential neural network.

\subsection{Sequential Neural Network}
\label{sec:sequential_neural_network}
Three sequential neural networks are used to analyze the sequence of features and perform rehabilitation exercise segmentation and counting. The core of all the networks is an LSTM trailed by a 1D CNN followed by a fully connected neural network.

\subsubsection{Many-to-many with Binary Sequence Output}
\label{sec:many_to_many_with_binary_output}
The sequential neural network is trained to output a binary sequence. Corresponding to the input at every timestamp (the body joints or extracted feature vector from every frame), the network generates one or zero. The output of one indicates that the input frame at a particular timestamp is a frame after a repetition of an exercise is complete or before the beginning of the next repetition, whereas an output of zero indicates that the input frame is during a repetition of an exercise. Using the generated output binary sequence, repetition segmentation is performed based on the occurrences of outputs of one among outputs of zero. Repetition counting is done by counting the number of segmented repetitions.

\subsubsection{Many-to-many with Density Map Output}
\label{sec:many_to_many_with_density_map_output}
The sequential neural network is trained to output a density map. A density map is a vector consisting of the same number of elements as the number of frames in the input, i.e., the number of timestamps in the sequential neural network. The ground-truth density maps are generated by approximating a Gaussian distribution between the start and end of each repetition \cite{hu2022transrac}. Repetition segmentation is performed by finding the peaks in the predicted density map. Repetition counting is done by counting the number of segmented repetitions.

\subsubsection{Many-to-one with Repetition Counts Output}
\label{sec:many_to_one}
By summing the outputs of the sequential neural network at consecutive timestamps, it will be converted to a many-to-one sequential neural network and is trained to output the number of repetitions. Unlike the two previous architectures, this architecture can only perform repetition counting and not segmentation.

The details of the parameters of the sequential neural networks are explained in Section \ref{sec:experimental_settings}.

\section{Experiments}
\label{sec:experiments}
This section evaluates the performance of the proposed method on three publicly available datasets using different evaluation metrics and in comparison to previous methods.

\subsection{Datasets}
\label{sec:datasets}

\subsubsection{KIMORE}
\label{sec:kimore}
The KIMORE dataset \cite{capecci2019kimore} contains RGB and depth videos along with body joint position and orientation data captured by the Kinect camera. The data were collected from 78 subjects, including 44 healthy subjects and 34 patients with motor dysfunction (stroke, Parkinson’s disease, and low back pain). Each data sample in this dataset is composed of one subject performing multiple repetitions of one of the five exercises: (1) lifting of the arms, (2) lateral tilt of the trunk with the arms in extension, (3) trunk rotation, (4) pelvis rotations on the transverse plane, and (5) squatting. The data samples were annotated in terms of exercise quality and technique.

The focus of this paper is on repetition segmentation and counting. KIMORE, however, does not contain such annotations. Two co-authors of this paper annotated the start and end of repetitions in each data sample using RGB video playback and the Aegisub software \cite{aegisub}. The annotations were verified by another two co-authors. The annotations were used as ground-truth labels for training and evaluation of neural network models. The annotations are available at https://github.com/abedicodes/repetition-segmentation. The mean (standard deviation) of the number of repetitions across the total 353 samples in the dataset is 4.70 (2.21).

In this paper, two modalities of data from the KIMORE dataset were used. See Section \ref{sec:methodology}. In the first setting, RGB videos were used as input data. The body joints were extracted by OpenPose \cite{pavllo20193d} and then inputted into the sequential models for analysis. In the second setting, the body joints captured by Kinect were directly considered as input to the sequential models.

\subsubsection{UI-PRMD}
\label{sec:uiprmd}
UI-PRMD \cite{vakanski2018data} is a dataset of physical therapy rehabilitation collected from 10 healthy individuals who performed 10 physical therapy rehabilitation exercises correctly and incorrectly to represent the performance of patients. Each data sample in this dataset is composed of one subject performing multiple repetitions of one of the 10 exercises, (1) deep squat, (2) hurdle step, (3) inline lunge, (4) side lunge, (5) sit to stand, (6) standing active straight leg raise, (7) standing shoulder abduction, (8) standing shoulder extension, (9) standing shoulder internal-external rotation, and (10) standing shoulder scaption. The annotations for repetition segmentation were available in UI-PRMD and were used in our experiments. The mean (standard deviation) of the number of repetitions across the total 200 samples in the dataset is 10.00 (0.00). This dataset contains body joints captured by Vicon and Kinect cameras. The Kinect data was used in our experiments.

\subsubsection{IntelliRehabDS}
\label{sec:intellirehabds}
IntelliRehabDS \cite{miron2021intellirehabds} is a dataset of body joints captured by the Kinect camera collected from 29 subjects, 15 patients and 14 healthy subjects. Each data sample in this dataset is composed of one subject performing multiple repetitions of one of the 9 physical rehabilitation exercises, (1) elbow flexion left, (2) elbow flexion right, (3) shoulder flexion left, (4) shoulder flexion right, (5) shoulder abduction left, (6) shoulder abduction right, (7) shoulder forward elevation, (8) side tap left, and (9) side tap right. The annotations for repetition segmentation were available in IntelliRehabDS and were used in our experiments. The mean (standard deviation) of the number of repetitions across the total 361 samples in the dataset is 5.69 (2.48).

\subsection{Evaluation Metrics}
\label{sec:evaluation_metrics}
In accordance with the literature, Off-By-One count Accuracy (OBOA) and MAE were used as evaluation metrics for repetition counting \cite{hu2022transrac,chunglong,hsu2021invariant}. Intersection-Over-Union (IOU) \cite{jiang2023stc} was used for repetition segmentation. A new metric for evaluating repetition segmentation was developed in this study, Mean Average Error in Frames (MAE-F). First, for every data sample, the average number of frames by which the start and end points of the predicted repetition deviate from those of the ground truth repetition is calculated. Then, these deviations are averaged over all data samples.

\subsection{Experimental Settings}
\label{sec:experimental_settings}
The dimension of the input to the sequential neural networks is determined based on the dimension of the feature vectors extracted from frames of the data samples. This dimension is 75, 43, and 118, when using the Kinect body joints, exercise-specific features \cite{guo2021exercise}, and their concatenation, respectively. This dimension is the number of neurons in the input layer of the LSTM. The number of neurons in the hidden layers of LSTM is set to two times the dimension of the feature vectors. In different exercises of different datasets, the number of LSTM layers varies from one to three layers.


For comparison, we implemented a \textit{modified} version of the TransRAC model \cite{hu2022transrac}, described in Section \ref{sec:related_work_video}. In place of using the encoder in TransRAC for feature extraction from multi-scale videos, multi-scale body joints were replaced. The other parts of the network remained unchanged and the network was trained from scratch to predict the density map. As a \textit{video-based} method, the pre-trained TransRAC on RepCount \textit{TransRAC} was also used to predict the density map for the RGB videos of exercises in the KIMORE dataset \cite{capecci2019kimore}.

Due to the fact that none of the datasets determined separate training and test sets, five-fold cross-validation was used. The loss function was a linear combination of the Kullback-Leibler divergence loss and L1 loss minimized by the Adam optimizer \cite{paszke2019pytorch}. The experiments were implemented in PyTorch \cite{paszke2019pytorch} and scikit-learn \cite{pedregosa2011scikit} on a server with 128 GB of RAM and NVIDIA 2080 Ti 12 GB GPU. The code of our implementation is available at https://github.com/abedicodes/repetition-segmentation.

\begin{table*}[]
\caption{(a) Mean absolute error and (b) off-by-one accuracy of repetition counting, and (c) intersection over union and (d) mean average error in frames of repetition segmentation on the KIMORE dataset \cite{capecci2019kimore} through five-fold cross-validation.
Density Map: many-to-many model with density map output,
General: One single model was trained and evaluated on all the samples in the dataset,
Pre-trained TransRAC: Pre-trained TransRAC model on RepCount video dataset [16],
Modified TransRAC: A modified version of the TransRAC model \cite{hu2022transrac} by removing the autoencoder feature extractor in TransRAC and replacing the body joints as the features,
Exercise Specific: Exercise-specific models were trained and evaluated on the samples of specific exercises in the dataset,
Density Map (LSTM Only): many-to-many model without CNN with density map output,
Binary sequence: many-to-many model with binary sequence output,
Counts: many-to-one model with repetition counts output.}
\begin{center}
\includegraphics[scale=.92]{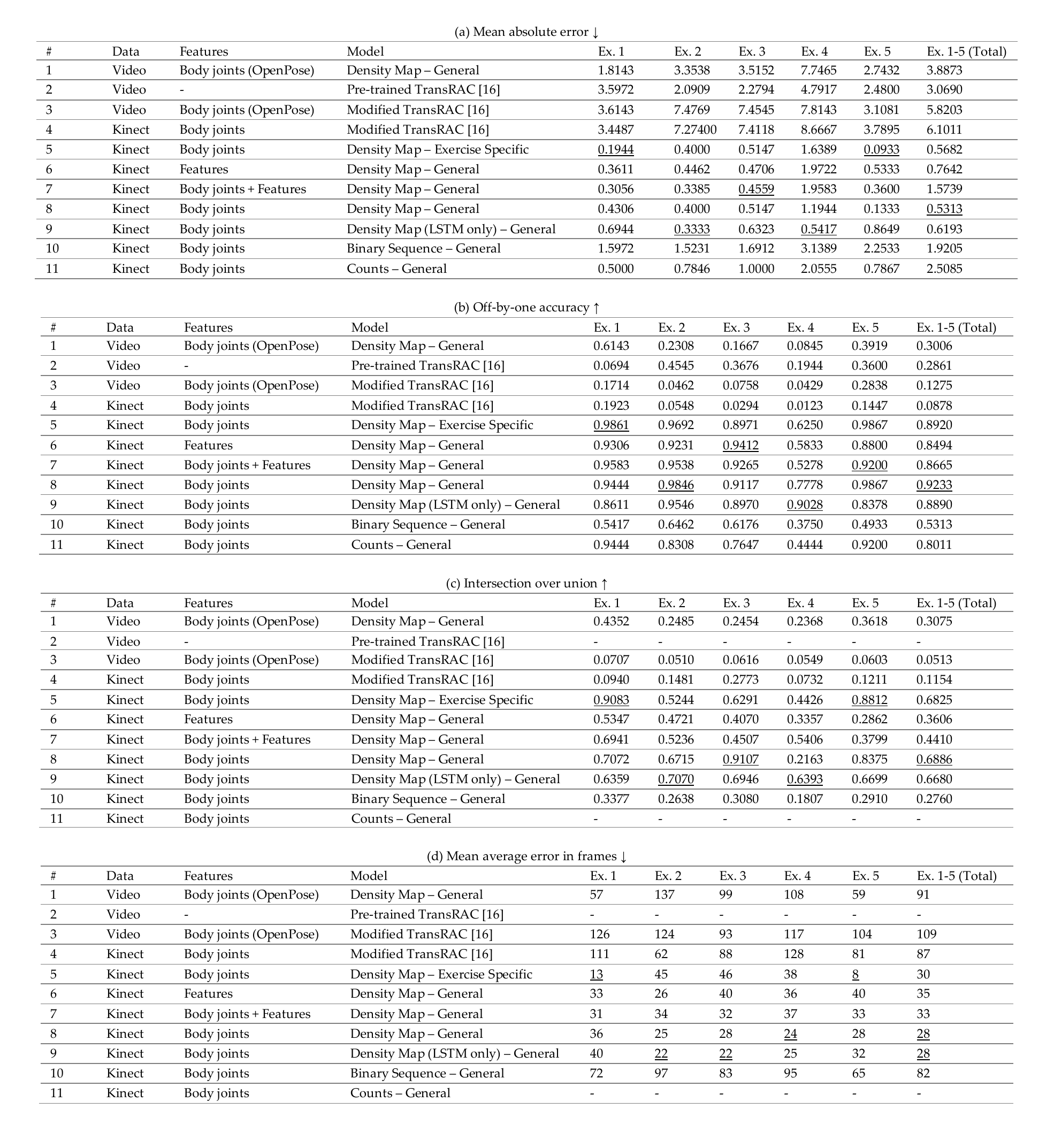}\\
\end{center}
\label{tab:kimore}
\end{table*}

\subsection{Experimental Results}
\label{sec:experimental_results}
\subsubsection{KIMORE}
\label{sec:kimore_results}
Tables \ref{tab:kimore} (a)-(b), and (c)-(d), respectively present the results of repetition counting, and segmentation on the KIMORE dataset \cite{capecci2019kimore} through five-fold cross-validation. The body joints captured by the Kinect camera in the dataset (rows \#4-11 in tables), the raw RGB videos in the dataset (\#2), and the body joints extracted from the RGB videos by OpenPose (\#1 and \#3) \cite{pavllo20193d} were analyzed in the experiments in Table \ref{tab:kimore}. It should be noted that repetition counting was carried out in rows \#1 and \#3-10 of Tables \ref{tab:kimore} (a)-(b) as a result of repetition segmentation. First, repetition segmentation was undertaken, and the number of repetitions was counted. The repetition counting was performed directly in rows \#2 and \#11 of Tables \ref{tab:kimore} (a)-(b).

The results of the many-to-many model with density map output (\#8) are superior to those of the many-to-many model with binary sequence output (\#10). The ground-truth density maps were generated by approximating a Gaussian distribution between the start and end of each repetition \cite{hu2022transrac}. The model that was trained on and predicted based on density maps was more robust to noise and fluctuations in body joints, as well as irregularities and imperfections in completing exercises in the patients in the KIMORE dataset.

There is a significant difference between the results of the two models described above (\#8 and \#10) and those of the many-to-one model with repetition counts output (\#11). One reason for this is that more ground-truth information was available for the training of the first two models. The first two models described above were provided with labels for every timestamp; however, the many-to-one model was provided with only one label for the entire network.

The results based on body joints captured by the Kinect camera (\#8) were far better than those extracted from RGB videos (\#1). Due to the low quality of videos in the KIMORE dataset, the high distance of the subjects to the camera, imperfect lighting conditions, and blurred faces in the dataset to preserve privacy, OpenPose \cite{pavllo20193d} had difficulty extracting body joints from the dataset, resulting in low-quality body joints compared to Kinect body joints. Therefore, the OpenPose \cite{pavllo20193d} input to the models was noisy, resulting in suboptimal performance.

Comparing the results of the many-to-many model with density map output (\#8) as described in Section \ref{sec:many_to_many_with_density_map_output} (an LSTM trailed by a 1D CNN and a linear layer) with the results of the many-to-many model with density map with only an LSTM trailed by a linear layer and without 1D CNN (\#9) shows the importance of including 1D CNN in the proposed model.

Across all settings, the proposed method outperformed the modified TransRAC method (\#3-4), Section \ref{sec:experimental_settings}. The reason for this can be attributed to two factors: the extreme complexity of the modified TransRAC model in comparison to the simple yet effective sequential models used in this paper, as well as the limited number of training samples in the KIMORE dataset. It is common for healthcare and rehabilitation datasets to have a small number of training samples. In this regard, it illustrates the necessity of taking the amount of data into account when selecting and designing the architecture of deep neural networks.

The RGB videos of exercises included in the KIMORE dataset were input into the original TransRAC model which was pre-trained on the RepCount video dataset \cite{hu2022transrac} (\#2). The results are much inferior to those predicted by the proposed architectures. The RepCount dataset contains videos of healthy subjects performing repetitive actions perfectly, while the KIMORE dataset contains rehabilitation exercises conducted by patients with imperfect repetitive actions.

Comparing the results of different features provided to the many-to-many model with density map output, body joints (\#8), features extracted from body joints (\#6), and their concatenation (\#7) indicates that for the repetition counting and segmentation tasks, there is no need to extract handcrafted features from body joints \cite{guo2021exercise} and the body joints as the input to the models result in the best performance.

According to the results of the general model trained and evaluated on all the samples in the dataset (\#8), as compared to the exercise-specific models trained and evaluated on specific exercises (\#5), exercise-specific models are not needed for repetition counting and segmentation tasks, and general models perform marginally better than exercise-specific models.

Comparing the performance of the models for specific exercises reveals that almost all methods had difficulty segmenting and counting repetitions for Ex. 4 in the KIMORE dataset. This is due to the fact that Ex. 4 involved pelvis rotations, i.e., movements along the z axis, which is relatively difficult to capture by depth cameras and analyze by models. As the movements in Ex. 4 differed from those in other exercises, which involved movements of the hands and feet along the x and y axes, it was difficult to generalize models to the samples in Ex. 4.

There is a similar overall trend in the performance of different methods in different settings across all four evaluation metrics, MAE, OBO, IOU, and MAE-F. Overall, the general many-to-many model with density map output trained with Kinect body joints achieved superior results with the lowest total MAE (0.5313) and the highest total OBO (0.9233) for repetition counting, as well as the highest total IOU (0.6886) and the lowest total MAE-F (28) for repetition segmentation.

\begin{table*}[]
\caption{Mean Absolute Error (MAE) and Off-By-One count accuracy (OBO) of repetition counting and Intersection-Over-Union (IOU) and Mean Average Error in Frames (MAE-F) of repetition segmentation on (a) the UI-PRMD \cite{vakanski2018data} and (b) the IntelliRehabDS \cite{miron2021intellirehabds} dataset through five-fold cross-validation.
Density Map – General: many-to-many model with density map output trained and evaluated on all the samples in the dataset,
Counts – General: many-to-one model with repetition counts output trained and evaluated on all the samples in the dataset.}
\begin{center}
\includegraphics[scale=1.]{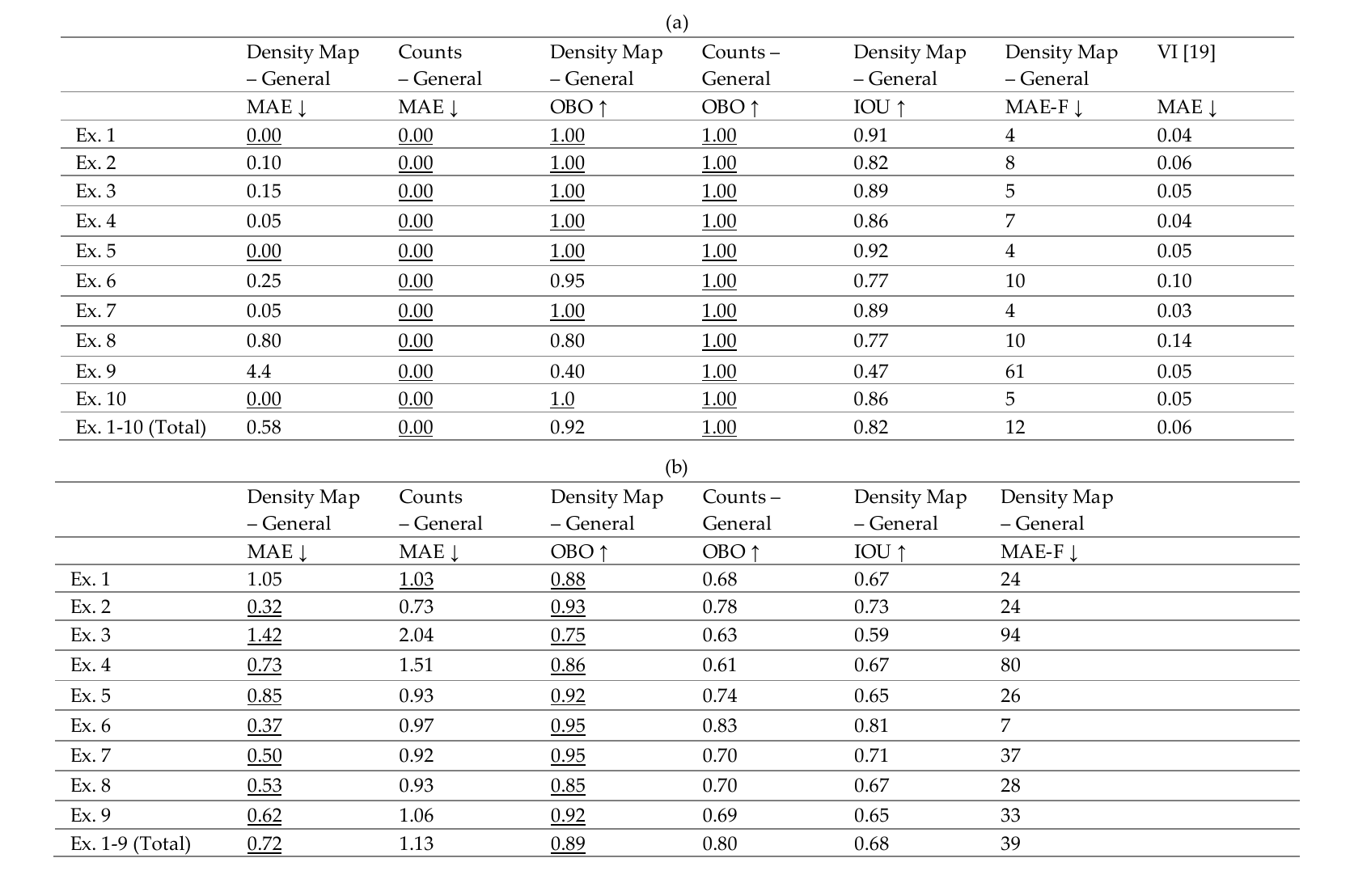}\\
\end{center}
\label{tab:ui-prmd-intellirehabds}
\end{table*}


\subsubsection{UI-PRMD}
\label{sec:uiprmd_results}
Table \ref{tab:ui-prmd-intellirehabds} (a) shows the results of repetition segmentation and counting on the UI-PRMD dataset \cite{vakanski2018data} using the general many-to-many model with density map output and the general many-to-one model with repetition counts output both trained and evaluated on Kinect body joints through five-fold cross-validation. Table \ref{tab:ui-prmd-intellirehabds} (a) illustrates that the repetition counting results of the many-to-many model with density map output are very similar to those of the many-to-one model with repetition counts output having zero errors in counting and outperforming the previous repetition counting method \cite{hsu2021invariant}. It is important to note that the number of repetitions across samples in the UI-PRMD dataset has a very low standard deviation. Using the proposed method, repetition segmentation is successfully performed with a total IOU of 0.82 and a total MAE-F of 12.

\subsubsection{IntelliRehabDS}
\label{sec:intellirehabds_results}
Table \ref{tab:ui-prmd-intellirehabds} (b) shows the results of repetition segmentation and counting on the IntelliRehabDS dataset \cite{miron2021intellirehabds} using the general many-to-many model with density map output and the general many-to-one model with repetition counts output both trained and evaluated on Kinect body joints through five-fold cross-validation. In Table \ref{tab:ui-prmd-intellirehabds} (b), the results of the many-to-many model with density map output are superior to those of the many-to-one model with repetition counts output. Using the proposed method, repetition segmentation is successfully performed with a total IOU of 0.68 and a total MAE-F of 39.

\begin{figure*}
    \centering
    \includegraphics[scale=.8]{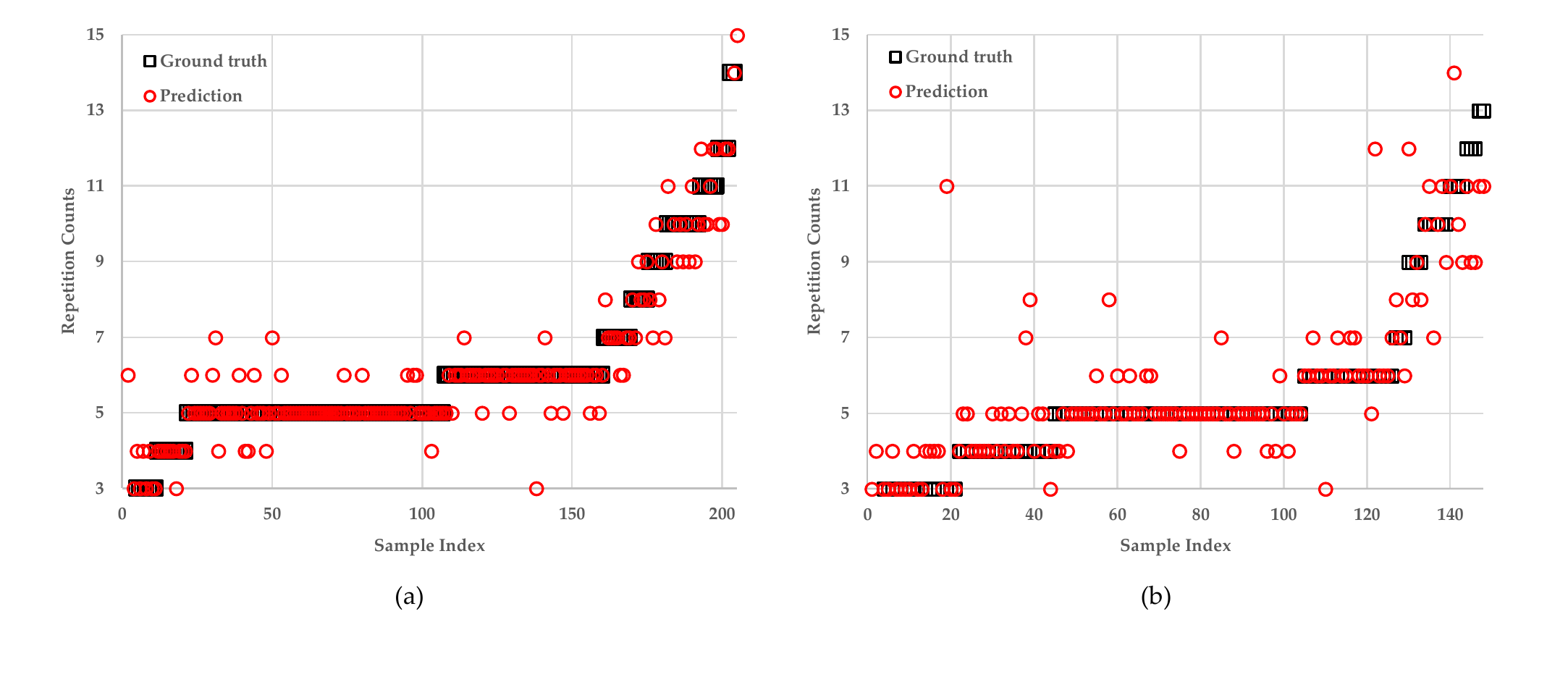}\\
    \caption{The ground truth values and predictions for repetition counts of the proposed many-to-many model with density map output on (a) healthy subjects and (b) patients in all the exercises in the KIMORE dataset \cite{capecci2019kimore}.}
    \label{fig:counts}
\end{figure*}

Figure \ref{fig:counts} illustrates the ground truth values and predictions for repetition counts of the proposed many-to-many model with density map output on (a) healthy subjects and (b) patients in all the exercises in the KIMORE dataset using five-fold cross-validation. The predictions closely follow ground-truth values in both populations, however, there are some deviations in patients as a result of irregularities in the duration and completion of exercise repetitions in patients. Figure \ref{fig:counts} (b) shows large deviations in predictions for samples with a large number of repetitions, such as samples with 12 and 13 repetitions. This can be attributed to the imbalanced distribution of samples in the dataset, i.e., having a small number of samples with large numbers of repetitions.

\section{Conclusion and Future Works}
\label{sec:conclusion}
The purpose of this study was to develop a learning-based method for segmenting and counting repetitions in rehabilitation exercises. On three publicly available rehabilitation exercise datasets, the proposed method successfully segmented and counted repetitions, and outperformed previous works, including a video-based method. This study represented the first work on repetition segmentation using skeletal body joints and data collected from patients, which is more challenging due to irregularities in exercise duration and completion. Despite being much lighter than video-based models \cite{hu2022transrac,chunglong,dwibedi2020counting,ferreira2021deep,cheng2023periodic,jacquelin2022periodicity,yu2021deep,zhang2021repetitive}, our sequential models require an initial stage of body-joint extraction from videos or directly by the use of depth cameras. Our body-joint-based method has the advantage of being more interpretable and capable of being used in a framework providing patients with actionable feedback on their exercises. The successful segmentation and counting of rehabilitation exercises using the proposed method is the first step in the development of an automated virtual rehabilitation platform capable of assessing exercise quality and providing feedback to patients and reports to clinicians. Future research may involve incorporating the attention mechanism \cite{10022210,nikpour2023spatio} into the current sequential models and extending the models to facilitate multi-task learning to perform exercise segmentation and assessment jointly.

\end{document}